\documentclass[twoside,11pt]{article}

\usepackage{mlhc}

\usepackage[usenames,dvipsnames]{xcolor}
\definecolor{shadecolor}{gray}{0.9}

\usepackage[english]{babel}
\usepackage[parfill]{parskip}
\usepackage{afterpage}
\usepackage{framed}

{\endMakeFramed}

\usepackage{lineno}

\usepackage{ragged2e}

\newcounter{parcount}

\usepackage{graphicx}
\usepackage[labelfont=bf]{caption}
\usepackage[format=hang]{subcaption}

\usepackage{booktabs}

\usepackage[algoruled]{algorithm2e}
\usepackage{listings}
\usepackage{fancyvrb}
\fvset{fontsize=\normalsize}

\hypersetup{citecolor=Blue}
\hypersetup{linkcolor=MidnightBlue}
\newcommand{\myfig}[1]{Figure~\ref{fig:#1}}

\usepackage
[acronym,smallcaps,nowarn,section,nonumberlist]{glossaries}%nogroupskip,
\glsdisablehyper{}
\lstdefinestyle{mystyle}{
    commentstyle=\color{OliveGreen},
    keywordstyle=\color{BurntOrange},
    numberstyle=\tiny\color{black!60},
    stringstyle=\color{MidnightBlue},
    basicstyle=\ttfamily,
    breakatwhitespace=false,
    breaklines=true,
    captionpos=b,
    keepspaces=true,
    numbers=left,
    numbersep=5pt,
    showspaces=false,
    showstringspaces=false,
    showtabs=false,
    tabsize=2
}
\newcommand{\g}{\, | \,}

\newcommand{\expfam}{\textsc{expfam}}

\newcommand{\DATA}{{\mathbf x}}
\newcommand{\data}[1]{x^{\textrm{#1}}}
\newcommand{\params}[1]{\beta^{\textrm{#1}}}
\newcommand{\PARAMS}{{\pmb \beta}}

\newcommand{\DEF}{\textrm{DEF}({\mathbf W})}

\usepackage{amsmath}

\ShortHeadings{Deep Survival Analysis}{Ranganath, Perotte, Elhadad and Blei}
\begin{document}

\title{Deep Survival Analysis}

\author{\name Rajesh Ranganath \email rajeshr@cs.princeton.edu \\
      \addr
       Princeton University \\
       Princeton, NJ 08540
       \AND
       \name Adler Perotte \email adler.perotte@columbia.edu  \\
       \addr
       Columbia University\\
       New York City, NY, 10032
       \AND
       \name No\'emie Elhadad \email noemie.elhadad@columbia.edu  \\
       \addr
       Columbia University\\
       New York City, NY, 10032
       \AND
       \name David Blei \email david.blei@columbia.edu \\
       \addr
       Columbia University\\
       New York City, NY, 10027
       } 

\maketitle

\begin{abstract}
The electronic health record (EHR) provides an unprecedented opportunity to
build actionable tools to support physicians at the point of care. 
In this paper, we introduce \emph{deep survival analysis}, a hierarchical generative approach to
survival analysis in the context of the EHR. 
It departs from previous approaches in two main ways: (1)
all observations, including covariates, are modeled jointly conditioned on a
rich latent structure; and (2) the observations are aligned by their failure
time, rather than by an arbitrary time zero as in
traditional survival analysis. Further, it handles heterogeneous
data types that occur in the EHR. 
We validate deep survival analysis by stratifying patients according to risk of
developing coronary heart disease (CHD) on
313,000 patients corresponding to 5.5 million months of observations.  When
compared to the clinically validated Framingham CHD risk score, deep survival analysis is
superior in stratifying patients according to their risk.
\end{abstract}

%!TEX root = 16mucmd.tex

\section{Introduction}

Our goal is to use electronic health record (EHR) data to estimate the
time of a future event of interest, namely, to carry out survival
analysis in a healthcare context.  Accurately estimating the time to an
event improves clinical decision support by allowing physicians to
take risk-calibrated actions. As a motivating example, consider coronary
heart disease (CHD). It is the leading cause of
death worldwide \citep{Hansson:2005, Pagidipati:2013}.  This condition,
also known as coronary artery disease or ischemic heart disease, is the
most common type of heart disease and causes 1 in every 4 deaths.  There
are effective preventative therapies for CHD that can significantly
reduce the risk of morbidity and mortality: antiplatelet
therapy~\citep{isis2:1988}, statin
therapy~\citep{4S:1994,Sackseta:1996,Shepherdetal:1995}, hypertensive
therapy~\citep{Nealetal:2000}, and lifestyle
interventions~\citep{Hjermann:1981, Schuler:1992}. 
Given the numerous effective strategies for primary prevention
(no prior CHD event) and secondary prevention (prior history of CHD
event), there is great value to identifying those individuals at high
risk of experiencing a CHD event. This is particularly important
because these interventions, albeit effective, are not risk
free.

The challenge of administering treatment based on risk pervades
the clinical decision process, and risk scores are in use
for many conditions, such as prostate cancer~\citep{Thompsonetal:2013},
breast cancer~\citep{gailetal:1989}, and stroke~\citep{Gageetal:2001}.

The standard approach to developing risk scores hinges on
using a curated set of patient data to regress
covariates to the time of failure. 
The significant covariates in the analysis are then summarized
in a easy-to-use table (for CHD see \citet{Wilsonetal:1998}). However, this
approach has serious limitations with respect to EHR
data.  First, regression requires complete
measurement of the covariates for all patients; in practice, many are
missing.  Second, the traditional approach requires all patients are
aligned based on some initial event (e.g., entry into trial, onset of a
disease related to event of interest, start of medication,
etc.); EHR data does not enjoy a natural alignment.
Third, the relationship between the
covariates and the time of the medical event is assumed to be linear,
possibly with some interaction terms; this limits the kind of 
relationships that may be found.

In this paper we propose a novel model for survival analysis from EHR
data, which we call \textit{deep survival analysis}.  Deep survival
analysis handles the biases and other inherent characteristics of EHR
data, and enables accurate risk scores for an event of interest.  The
key contributions of this work are:
\begin{itemize}

\item Deep survival analysis models covariates and survival time in a
  Bayesian framework.  This simplifies working with the missing
  covariates prevalent in the EHR.

\item Deep exponential families~\citep{Ranganath:2015}, a deep latent
  variable model, forms the backbone of the generative process. This
  results in a non-linear latent structure that captures complex
  dependencies between the covariates and the failure time.

\item Rather than enforcing an artificial time zero alignment for all
  patients, deep survival analysis aligns all patients by their
  failure time (i.e., the event occurs or data is right censored).

\item Good preprocessing of EHR data allows deep survival analysis to
  include heterogeneous data types. In our study, we include vitals, laboratory
  measurements, medications, and diagnosis codes.
\end{itemize}
We studied a large dataset of 313,000 patient records and used deep
survival analysis to assess the risk of coronary heart disease. 
Deep survival analysis better stratifies patients than
the gold-standard, clinically validated CHD risk
score~\citep{Wilsonetal:1998}.

This paper is organized as follows. Section~\ref{sec:survival} reviews
the fundamentals of traditional survival analysis and motivates the
need for better modeling tools for EHR data. Section~\ref{sec:deep}
reviews deep exponential families~\citep{Ranganath:2015} and
Section~\ref{sec:alignment} discusses our alignment strategy for deep
survival analysis. Section~\ref{sec:generative} describes the modeling
assumptions behind deep survival analysis;
Section~\ref{sec:inference} gives details of our scalable variational inference
algorithm. 
Section~\ref{sec:experiments} describes the clinical
scenario of CHD, data, experimental setup, baseline, and evaluation
metrics.  Finally, Sections~\ref{sec:results} and~\ref{sec:concl}
discuss our results and conclusions.

%%% Local Variables:
%%% mode: latex
%%% TeX-master: "16mucmd"
%%% End:

%!TEX root = 16mucmd.tex

\vspace*{-1ex}
\section{Survival Analysis}\label{sec:survival}
\vspace*{-1ex}

In this section, we provide background on the task of survival
analysis. We review the traditional approaches, along with several
variants that are relevant to our work. We
then delve into two of the primary limitations of current survival analysis
techniques, which hinder their use in EHR data.

\vspace*{-1ex}
\subsection{Traditional Survival Analysis}

Survival analysis models the time to an event from a common
start~\citep{Kaplan:1958}. Examples of survival data include time to
delivery from conception and time to retirement from birth. 
Survival observations consist of
two varieties. The first are observations for which
the exact failure time is known. The second, called
censored observations, are observations for which the failure 
time is known to be greater than a particular time. Both types
 are represented as $(t, c)$, pairs of positive times
and binary censoring status.

Survival modeling assumes the observations, both censored and uncensored, come from
an unknown distribution. The two traditional methods for estimating
the survival distribution are the Kaplan-Meier estimator~\citep{Kaplan:1958} and
Cox proportional hazards~\citep{Cox:1972}. The Kaplan-Meier estimator
forms a nonparametric estimator for the survival function, one minus
the cumulative distribution function. Intuitively, it breaks a 
stick of length one at points proportional to the fraction of 
patients that survived until that point.
Cox proportional hazards generalizes this estimator to include covariates. 
Finally, Bayesian variations of these 
methods place priors on the parameters~\citep{Hjort:1990}.

\begin{figure*}[t]
\centering
\includegraphics[width=1\textwidth]{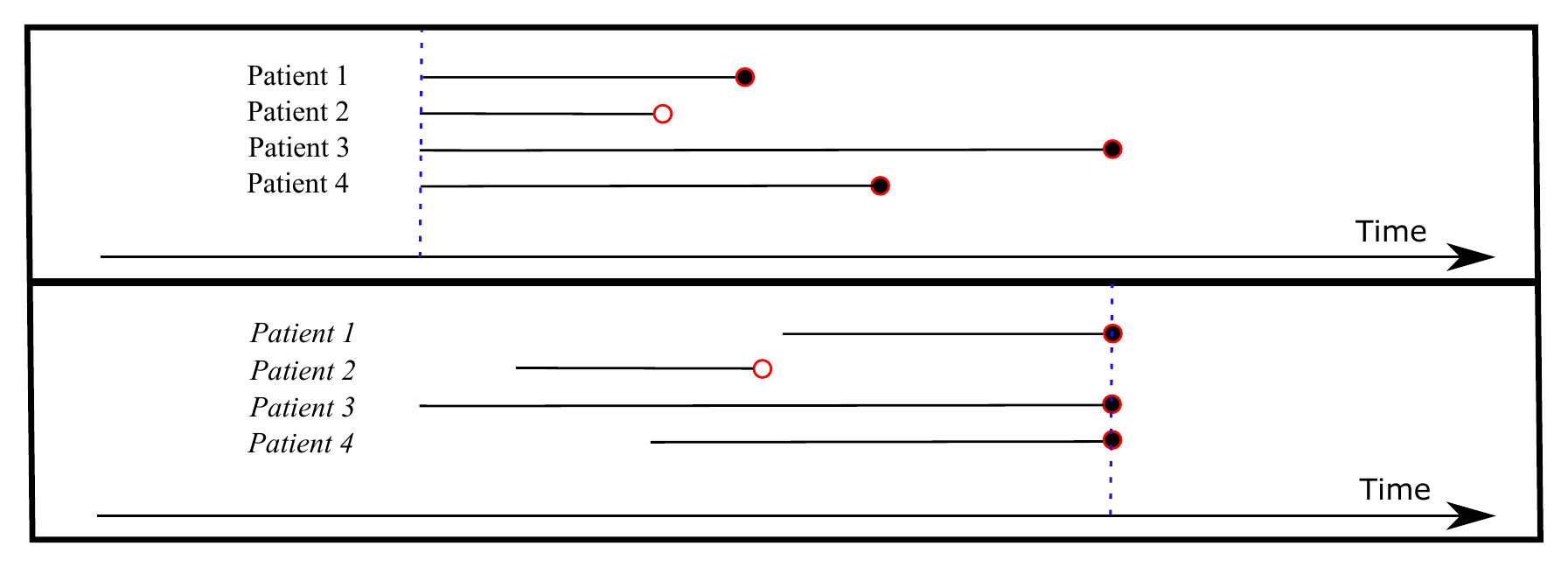}
\vspace*{-3ex}
\caption{A comparison of traditional survival analysis (top frame) and
  failure aligned survival analysis (bottom frame).  A filled circle represents an
  observed event, while an empty circle represents a censored one.
  In the case of standard
  survival analysis, patients in a cohort are aligned by a starting event. In
  failure aligned survival analysis, patients are aligned by a failure event.
  \label{fig:alignment}
}
\end{figure*}

\subsection{Limitations of Traditional Survival Analysis for EHR
Data}\label{sec:survival_ehr} 

There are three significant limitations to using traditional
survival methods on EHR data.

First, EHR data is usually high-dimensional and very
sparse~\citep{hripcsak2013next, pivovarov2014identifying}. 
This makes it difficult to use traditional conditional models, 
which cannot easily handle missing covariates

Second, traditional methods require aligning all patients based
on a synchronization event. This event can be entry to a clinical
trial, the date of an intervention, or the onset of a
condition. However, the EHR for a patient can begin at any point
in their lifetime and at any point in their disease progression. Thus
careful definition of entry point into study are required
when experimenting with traditional survival techniques on  EHR data
(e.g., ~\cite{hagar2014survival,15perotte_ckd}).  
In this work, we seek methods that are able to evaluate risk
at any point in time, not only points in time that correspond to such
a synchronization event.

Third, regression-based approaches to survival analysis often assume a
linear function of the covariates. Nonlinear interaction terms are
sometimes introduced but must be limited (often based on expert
opinion) because of the combinatorial explosion of possibilities. Models with greater
flexibility can incorporate nonlinear relationships between
combinations of covariates and time-to-event.

%!TEX root = 16mucmd.tex

\section{Deep Survival Analysis}

We introduce \emph{deep survival analysis}, a hierarchical generative approach
for survival analysis. It departs from previous approaches in two primary ways.
First, all observations, including covariates, are modeled jointly and conditioned
on a rich latent structure. Second,
patient records align by their failure time rather than by entry time,
thus resolving the ambiguity of entry to the EHR.

\subsection{Deep Exponential Families}\label{sec:deep}
When not modeling the covariates, missing
covariates are usually imputed using population-level statistics. In
contrast, we build a joint model for both the covariates and the survival
times, where the covariates and survival times are specified conditioned on a
latent process. This strategy requires a rich latent process; we use deep
exponential families~\citep{Ranganath:2015}.

Deep exponential families (DEF) are a class of multi-layer probability models
built from exponential families~\citep{Brown:1986}. In deep exponential
families, each observation has $L$ layers of latent variables.
Each layer conditions on the previous layer of latent variables. Formally, let
$n$ be the index of the data, $\expfam(\cdot)$ be an exponential family
distribution with natural parameter, $g(\cdot; W)$ 
denote a link function with parameters $W$ with prior $p(W)$, and
$\eta$ be a hyperparameter. The generative process for the latent
variables is
\begin{align*}
z_{n,L,j} &\sim \expfam_L(\eta)
\\
z_{n, \ell, j} &\sim \expfam_\ell(g_\ell(z_{n, l+1}, W_{l, j})).
\end{align*}
The observations for the $i$th data point are drawn conditional
on the vector $z_{i, 1}$. We shorthand the draw of the last layer, $z_{n, 1}$ as 
$z_n \sim \DEF$. DEFs have been successful at modeling
text, recommender systems, and images. They handle missing
data better than competitive latent variable models and state-of-the-art
density estimators~\citep{Rezende:2014,Ranganath:2015}. Thus, they are a
promising prior for the latent structure to model survival in the EHR.

\subsection{Alignment by Failure}\label{sec:alignment}
Censored survival observations are pairs $(t_i, c_i)$,
where $t_i$ denotes the time of the $i$th observation, 
and $c_i$ marks whether failure or censoring occurs at that time.
Traditionally, the time $t_i$ is measured from a common start point for each
observation, such as birth or pregnancy. As we mentioned earlier, this type of
alignment is inappropriate in the context of the EHR because people enter the EHR
in different ways relative to to their underlying health. This limitation
of survival analysis was acknowledged
by~\citet{mccullagh2013survival}, but their solution has only been tested on a small
dataset, and does not directly apply to censored observations. 

We consider an event-centric ordering, which measures time backwards
from the event of interest, rather than measuring time forward from an
artificial start time. At the event all patients share the defining
characteristics of the event, thus patients are similar at time zero
under this alignment. We handle censored observations as interval observations. 

For each time point, this alignment models the time to failure
from that time point. This is
a positive number which decreases when approaching failure.
Censored events differs. For censored patients, their failure
must happen (if it happens) after their last interaction with the EHR;
their time of interest is an interval greater than the time of their
last visit to the EHR.
As such, event-centric ordering of data consists of pairs $(t_i, c_i)$. 
See~\myfig{alignment} for a graphical 
comparison of this versus the standard survival setup.

In this approach, every interaction with the EHR has a (possibly censored) time
from event/failure associated with it. This means the interactions can be modeled
exchangeably, which trades statistical efficiency of persisting patient
information over time for computational efficiency. We take this approach.
Each event of interest in the EHR represents a different
survival alignment frame. Thus, we can investigate several survival
tasks by choosing an event of interest and aligning by its timing.
This contrasts traditional survival analysis, which requires a careful
decision about the start time.

To model the time from event, we use a Weibull distribution, a popular
distribution in survival analysis.  Let $\lambda$ be the scale and $k$ be the
shape, the Weibull distribution is
\begin{align*}
p(t) = k / \lambda \left(t / \lambda \right)^{k-1} e^{-\left(\frac{t}{\lambda}\right)^k}.
\end{align*}
It has support over the positive reals and its parameters are constrained to be 
positive. Its expectation is $\lambda \Gamma(1 + \frac{1}{k})$. The parameter $k$ control how the density looks. When $k < 1$ most of the mass is concentrated 
near zero; when $k=1$ this distributions matches the exponential; when $k>1$, the distribution places mass around its expectation. For a censored observation, 
The likelihood is the amount of probability the model
places after censoring, i.e., one minus the cumulative
distribution function.
For the Weibull, this is
$\exp(-\left(\frac{t}{\lambda}\right)^k)$, which gets large as the scale grows.

\subsection{Generative Process for Deep Survival Analysis}\label{sec:generative}
Let $\DATA$ denote the set of covariates, $\PARAMS$ be the parameters for the data with some
prior $p(\PARAMS)$, $k$ be a fixed scalar, and let $n$ index an
observation. The generative model for deep survival analysis is 
\begin{align}
b &\sim \textrm{Normal}(0, \sigma_b)
\nonumber \\
a &\sim \textrm{Normal}(0, \sigma_W)
\nonumber \\
z_n &\sim \DEF
\nonumber \\
\DATA_n &\sim p(\cdot \g \PARAMS, z_n)
\nonumber \\
t_n &\sim \textrm{Weibull}(\log(1 + \exp(z_n^\top a  + b), k).
\label{eq:model}
\end{align}
The latent variable $z_i$ comes from a DEF which then generates
the observed covariates and the time to failure. The function $\log(1 + \exp(\cdot))$,
called the softplus, maps from the reals to the positives to output a valid scale for the Weibull. Given covariates $\DATA$, the model makes predictions via the 
posterior predictive distribution:
\begin{align*}
p(t \g \DATA) = \int_z p(t \g z) p(z \g \DATA) dz.
\end{align*}
The complexity of the predictions depends on the complexity
of the distribution $z$. Note this predictive distribution exists and is consistent
even if data are missing. 

For electronic health records the $\DATA$ contain several
data types. We consider laboratory test values (labs), medications (meds),
diagnosis codes, and vitals. We assume each of these data types are generated
independently, conditional on the latent structure
\begin{align*}
p(\DATA_n \g \PARAMS, z_n) = p(\data{labs}_n \g z_n, \params{labs}) p(\data{meds}_n \g z_n, \params{meds}) p(\data{diagnoses}_n \g z_n, \params{diagnoses}) p(\data{vitals}_n \g z_n, \params{vitals}).
\end{align*}
We emphasizes that they are marginally dependent.

The data types in the EHR can be grouped by whether they are real valued (labs
and vitals) or counts (diagnoses and medications). Next we define the
likelihood for each group.

\paragraph{Real-Valued Observations.}
Real-valued observations in EHR are heavy tailed and are
prone to data entry errors~\citep{hauskrecht2013outlier}. This leads to extreme
outliers that may badly corrupt estimates of non-robust models such as those
based on the Gaussian. We model the real-valued data with the
Student-t distribution, a continuous mixture of Gaussians across scales, which
is more robust to outliers.
Given parameters,
$\params{labs}_{W_i}, \params{labs}_{b_i}$ and degrees of freedom $\nu$, the conditional density of the $i$th lab, is 
\begin{align*}
p(\data{labs}_{n, i} \g \params{labs}_{W_i}, \params{labs}_{b_i}, z_n) = \frac{\Gamma(\frac{\nu + 1}{2})}{\sqrt{\nu \pi}} \left(1 + \frac{(\data{labs}_{n, i} - (z_n^\top {\params{labs}_{W_i}} + \params{labs}_{b_i})   ^2}{\nu}\right)^{-\frac{\nu + 1}{2}}.
\end{align*}
This is a Student-t distribution whose mode is at $z_n^\top
{\params{labs}_{W_i}} + \params{labs}_{b_i}$ a function of both the data point
specific latent variables and the parameters shared  across data points. The
degrees of freedom controls to which extent the distribution resembles a
Naussian, where large values look more Gaussian. We place Gaussian priors on
both $W_i$ and $b_i$. The likelihood follows similarly for the vitals.

\paragraph{Count Valued Observations.}
Unlike the laboratory tests and the vitals, the count-valued observations
are highly dimensional and sparse. To handle counts robustly, we model them as
binary values, one if the count is non-zero and zero
otherwise. We model the $i$th medication with parameters $\params{meds}_{W_i}$
as  \begin{align*}
p(\data{meds}_{n, i} \g \params{meds}_{W_i}, z_n) \sim \textrm{Bernoulli}(1 - \exp(z_n^\top \params{meds}_{W_i}))),
\end{align*}
where $\params{meds}_{W_i}$ has a log-Gaussian prior. This likelihood has the
added benefit that the total likelihood and its gradient can be
computed in time proportional to the number of nonzero
elements~\citep{Ranganath:2015b}. We overdisperse the Bernoulli likelihood by
the number of medications to balance this component with the time from failure.
The diagnoses are modeled in the same manner. 

%!TEX root = 16mucmd.tex

\section{Experimental Setup}\label{sec:experiments}

We apply deep survival analysis to data from a large metropolitan
hospital. We use the fitted model to predict coronary heart disease risk (CHD). Access to
data and experiments was approved after review by the Columbia University
Institutional Review Board.

\subsection{Dataset}

Our dataset comprises the longitudinal records of 313,000 patients
from the Columbia University Medical Center clinical data warehouse. The
patient population included all adults (\textgreater 18 years old) that
have at least 5 months (not necessarily consecutive) where at least one
observation was recorded.

The patient records contain documentation resulting from all settings,
including inpatient, outpatient, and emergency department visits. Observations
included 9 vital signs, 79 laboratory test measurements, 5,262 
medication orders, and 13,153 diagnosis codes. 

\vspace*{-1ex}
\paragraph{Data Preprocessing.}
All real-valued measurements and discrete variables were aggregated at the
month level, leading to binned observations for each patient and
for each month the patient had any recorded observation. The expected value
over the course of each month was computed for continuous measurements such as
vitals and laboratory measurements and the presence of discrete
elements such as medication orders and diagnosis codes was encoded as
a binary variable.

\subsection{Baseline and Model Setup}

\vspace*{-1ex}
\paragraph{Baseline.}
The Framingham CHD risk score was developed in 1998 and is one of the
earliest validated clinical risk scores. It is a gender-stratified
algorithm for estimating the 10-year coronary heart disease risk of an
individual. Aside from gender, this score takes into consideration
age, sex, LDL cholesterol, HDL cholesterol, blood pressure, diabetes,
and smoking.  For example, a 43-year-old (1 point) male patient with
an LDL level of 170 mg/dl (1 point), an HDL level of 43 mg/dl (1 point),
a blood pressure of 140/90 (2 points), and no history of diabetes (0
points) or smoking (0 points) would have a risk score of 5 which would
correspond to a 10 year CHD risk of 9\%~\citep{Wilsonetal:1998}.

The score was validated using curated data from the Framingham Heart Study. It
was shown to have good predictive power of 10-year risk  with a concordance of
0.73 for men and 0.77 for women. However, this score has lower performance
when applied to EHR data~\citep{Pikeetal:2016}.

\vspace*{-1ex}
\paragraph{Model Setup and Hyperparameters.}
We set the shape of the Weibull to be 2. The exponential family used
inside the DEF is a Gaussian. The mean and inverse softplus variance
functions for each layer are a 2 layer perceptron with rectified
linear activations. We set Normal priors to have mean zero and
variance one.

We let all methods run for 6,000 iterations and assess convergence on a 
validation set.  On a 40-core Xeon Server with 384 GB of RAM, 6,000
iterations for all patients in the training set completed in 7.5
hours. Due to the high variance in patient record lengths, we subsample
observations during inference inversely to the
length of the patient records. 

%!TEX root = 16mucmd.tex

\vspace*{-1ex}
\paragraph{Inference.}\label{sec:inference}
We approximate the posterior distribution
with variational inference~\citep{Jordan:1999} for the observation specific latent
variables in the DEF and do maximum-a-posteriori inference on the parameters.
We choose the approximating family to be the mean-field family where each latent
variables gets its own independent parameterization. We use black box variational 
methods~\citep{Ranganath:2014} with reparameterization gradients~\citep{Kingma:2014,Rezende:2014} to approximate the posterior without needing model specific computation. 
To scale to the large data,
we subsample data batches~\citep{Hoffman:2013} of size 240 
and parallelize computation across
data in a batch. We use RMSProp with scale $0.0001$ and Nesterov
momentum of $0.9$ as learning rates during optimization.

\vspace*{-1ex}
\subsection{Evaluation}
\vspace*{-1ex}

Of the 313,000 patients in the study, 263,000 were randomly selected
for training, 25,000 for validation, and 25,000 for testing. We
assessed convergence with the validation cohort and evaluated
concordance on the test cohort. A CHD event was defined as the
documentation of any ICD-9 diagnosis code with the following prefixes: 413
(angina pectoris), 410 (myocardial infarction), or 411 (coronary
insufficiency). In our experiments, we vary the dimensionality of
$z_n$ to assume the values of K $\in \{5,10,25,75,100\}$. The layer
size for the perceptrons were set to equal the dimensionality of $z_n$
in each experiment. We evaluate both the baseline risk score and
deep survival analysis with the
concordance~\citep{Harrell:1982}.

While concordance enables the comparison of deep survival analysis
to the baseline, it only roughly captures the accuracy of the temporal
prediction of the models. In deep survival analysis, we can
compute the predictive likelihood of the held-out set according to the
model, which enables us to capture how well the model predicts failure in time.
For internal model validation, we thus rely on predictive likelihood. 
Predictive likelihood is evaluated as the expected log probability of
the observed time until failure
conditioned on the observed covariates for a given patient in a given
month.

%!TEX root = 16mucmd.tex

\section{Results}\label{sec:results}

Missing data is a core challenge of 
electronic health record data analysis and temporal analysis. We  
first report the extent of incomplete observations in our dataset. 
We then report results for the baseline CHD risk model
and deep survival analysis. We also report internal validation of deep 
survival analysis which include only a single data type.

\subsection{Missing observations in EHR data}
For estimating CHD risk in 10 years, the widely used guideline-based CHD risk
score calculators used routinely by clinicians require input of seven
variables: age, sex, current smoking status, total cholesterol level, HDL
cholesterol level, systolic blood pressure, and whether patient takes blood
pressure medication. We examined how many patients had at least one month in
their record, where the most basic, critical set of variables were observed
(LDL level, HDL level, and blood pressure). In the full dataset, only
$11.8\%$ of patients have a complete month, and $1.4\%$ of months are complete.

\begin{table}[htbp]
  \centering 
  \begin{tabular}{ll}
  \toprule
    Model & Concordance (\%) \\ \midrule
    \textbf{Baseline Framingham Risk Score} & \textbf{65.57} \\ 
    Deep Survival Analysis; K=10 & 69.35  \\ 
    Deep Survival Analysis; K=5  & 70.45 \\ 
    Deep Survival Analysis; K=25 & 71.20 \\ 
    Deep Survival Analysis; K=75 & 71.65 \\ 
    Deep Survival Analysis; K=100 & 72.71 \\
    \textbf{Deep Survival Analysis; K=50} &  \textbf{73.11} \\ 
    \bottomrule
  \end{tabular}
  \caption{Concordance on a held-out set of 25,000 patients for different values of K and for the baseline risk score. All deep survival analysis dimensionalities outperform the baseline.} 
  \label{tab:results}
\end{table}

\subsection{Model Performance and Predictive Likelihood}

The baseline CHD risk score yielded 65.57\% in concordance over the held out test
set. Table~\ref{tab:results} shows the concordance of the deep survival
analysis for different values of K. When considering the full deep
survival with all data types considered, the best performance was obtained
for K=50.

When examining the deep survival analysis with the best concordance on the held
out set (K=50), we then asked how well each individual data type predicts
failure. The following four models were thus trained: deep survival analysis 
including vitals only, diagnosis codes only, laboratory tests only, and
medications only. All models included age and gender. Their individual
predictive likelihood was computed on the same month bins, even in the absence
of observations of a specific data type. The diagnosis-only model yielded the
best predictive likelihood.

\begin{table}[h]
  \centering 
  \begin{tabular}{ll} 
  \toprule
   Data Type & Likelihood \\ \midrule
    Medications Only &  -1.24899 \\
    Laboratory Tests Only & -0.998774 \\ 
    Vitals Only &  -0.961827 \\ 
    Diagnoses Only & -0.855385 \\
    \bottomrule
  \end{tabular}
  \label{tab:types} 
  \caption{Predictive likelihood of deep survival analysis (K=50) for
  individual data types. The diagnoses perform best.}
\end{table}

%!TEX root = 16mucmd.tex

\section{Discussion}\label{sec:concl}

In this paper we introduce a new method for survival analysis built to handle
the inherent characteristics of EHR data. While traditional survival analysis
requires carefully curated research datasets, our approach easily handles
the sparsity and heterogeneity of EHR observations. We estimate deep survival analysis on the entire data from a large metropolitan
hospital in a matter of hours. When compared to one of the state-of-the-art, clinically
validated risk score in the context of coronary heart disease, deep survival analysis yields
a more accurate stratification of patients. 
Our approach holds particular promise for developing risk scores from
observational data for conditions where there is no known risk score.

\pagebreak

\acks{This work is supported by NSF \#1344668, NSF IIS-1247664, ONR N00014-11-1-0651, DARPA
FA8750-14-2-0009, DARPA N66001-15-C-4032, Adobe, The Sloan Foundation, The Seibel Foundation, and The Porter Ogden Jacobus Fellowship}

\bibliography{16mucmd}

\end{document}